\def\year{2022}\relax
\documentclass[letterpaper]{article} 
\usepackage{aaai22}  
\usepackage{times}  
\usepackage{helvet}  
\usepackage{courier}  
\usepackage[hyphens]{url}  
\usepackage{graphicx} 
\usepackage{natbib}  
\usepackage{caption} 
\DeclareCaptionStyle{ruled}{labelfont=normalfont,labelsep=colon,strut=off} 
\usepackage{algorithm}
\usepackage{algorithmic}
\usepackage{amsfonts,amssymb}
\usepackage{textcomp}
\usepackage{amsmath}

\usepackage{newfloat}
\usepackage{listings}
\floatstyle{ruled}
\newfloat{listing}{tb}{lst}{}
\floatname{listing}{Listing}
\title{\emph{RRL}: Regional Rotation Layer in Convolutional Neural Networks}
\author{
    Zongbo Hao,
    Tao Zhang,
    Mingwang Chen,
    Kaixu Zhou    
}
\affiliations{

    University of Electronic Science and Technology of China\\
    No.4, Section 2, North Jianshe Road\\
    Chengdu, China, 610054\\
    zbhao@uestc.edu.cn, zhangtao897476472@126.com, chenmingwang815@163.com, 1241510971@qq.com
    
}

\usepackage{bibentry}

\begin{document}

\maketitle

\begin{abstract}
Convolutional Neural Networks (CNNs) perform very well in image classification and object detection in recent years, but even the most advanced models have limited rotation invariance. Known solutions include the enhancement of training data and the increase of rotation invariance by globally merging the rotation equivariant features. These methods either increase the workload of training or increase the number of model parameters. To address this problem, this paper proposes a module that can be inserted into the existing networks, and directly incorporates the rotation invariance into the feature extraction layers of the CNNs. This module does not have learnable parameters and will not increase the complexity of the model. At the same time, only by training the upright data, it can perform well on the rotated testing set. These advantages will be suitable for fields such as biomedicine and astronomy where it is difficult to obtain upright samples or the target has no directionality. Evaluate our module with LeNet-5, ResNet-18 and tiny-yolov3, we get impressive results.
\end{abstract}

\section{Introduction}

Deep learning and convolutional neural networks have made great progress in many tasks such as image classification and object detection. The inherent properties of convolution and pooling layer alleviate the influence of local translation and distortion. However, due to the lack of the ability to process large rotation of image, convolution neural networks are limited in some visual tasks, including target boundary detection \cite{m:16,d:05}, multi-directional target detection \cite{c:16} and image classification \cite{j:15,l:16}. In recent years, CNN based image classification and object detection have been used in biomedical, industrial and astronomical research. In these fields, objects can appear in any direction, such as microscopic images, objects on conveyor belts or objects observed. So, the research on rotation invariance of neural networks has been more and more important.

At present, most deep networks use data augmentation to make the network recognize objects in different directions \cite{o:96,q:18,t:19,ma:16}, or merging the rotation equivariant features \cite{g:19,w:20}. These methods either increase the workload of training or increase the number of model parameters. In this paper, by making the feature maps before and after convolution rotation equivariant, the whole neural network is rotation invariant. With rotation angle $\theta\in\{0,90,180,270\}^{\circ}$,  the feature maps are completely the same. When the input image is rotated with arbitrary angle, there is only small difference between feature maps before and after rotation.

The main contributions of this paper are:

1) A Local Binary Pattern(\emph{LBP}) operator based Regional Rotation Layer (\emph{RRL}) is proposed. \emph{RRL} can be embedded in CNNs, without the need for substantial changes to the network to achieve rotation invariant. 

2) Without learning new parameters, \emph{RRL} makes the feature maps before and after convolution satisfy the rotation equivariance, and thus makes the entire neural network rotation invariant. With rotation angles\ $\theta\in\{0,90,180,270\}^{\circ}$, the feature maps are exactly the same. With arbitrary rotation angle, there is a small distinction between feature maps.

3) Evaluate \emph{RRL} with LeNet-5, ResNet-18 and tiny-yolov3, we get impressive results.

\section{Related Work}

For deep learning based methods, the most direct way is data augmentation \cite{v:01}, which simply changes the size and direction of the input images to create more training data. TI-Pooling \cite{l:16} uses the rotated image as input, and applies a pooling layer before outputting features to unify the network's results for different rotation angles. Dieleman proposed a deep neural network model that uses translational and rotational symmetry to classify galaxy morphology \cite{d:15r}. They create multiple rotated and flipped galaxy image samples, and then concatenate the feature maps to the classifier. Polar Transformer Networks (PTN) \cite{e:18} converts the input to polar coordinates. PTN is composed of a polar coordinate prediction module, a polar coordinate conversion module, and a classifier to achieve translation invariance and equal changes in expansion/rotation groups. \cite{j:19} also proposed a Polar Coordinate Convolutional Neural Network (PCCNN) to convert the input image to polar coordinates to achieve rotation invariance. The overall structure of the model is the same as the traditional CNN, except that the central loss function is used to learn rotation invariant features. In addition, \cite{c:16g} proposed Group Equivariant Convolutional Networks (GCNN) as a special case of controllable CNN, which proved that the spatial transformation of the image can be corresponded in the feature map and the filter. GCNN is composed of group convolution kernels. These convolutions include filter rotation and merging operations on the rotation. Group convolution is limited to integer multiples of 90\textdegree rotation and flipping. Cohen et al. also proposed steerable CNNs \cite{co:16steerable} and Spherical CNNs \cite{co:18spherical} to achieve rotation equivariant. Steerable CNNs are limited to discrete groups, such as discrete rotations acting on planar images or permutations acting on point clouds. Spherical CNNs show good robustness. Because FFT and IFFT are used in spherical convolution, some information will be lost in the conversion process. Spherical convolution achieves rotation invariance for ideal 3D objects, and there is no interference of background or other noise. If there are multiple 3D objects in the natural scene, the 3D objects must be segmented first, and then the rotation invariant features are extracted. The Rotation Equivariant Vector Field Networks (RotEqNet) \cite{m:17} uses multiple rotation instances of a uniform standard filter to perform convolution, that is, the filter is rotated at different intervals. Although the RotEqNet model is small, the increasing in the number of convolution kernels brings more memory and longer computing time. \cite{d:16} encodes cyclic symmetry in CNNs by parameter sharing to achieve rotation equivariant. They introduce four operations: slice, pool, stack and roll. The operations can be cast as layers in a neural network, and build networks that are equivariant to cyclic rotations and share parameters across different orientations. But the operations change the size of the minibatch (slicing, pooling), the number of feature maps (rolling), or both (stacking). To alleviate the excessive time-consuming and memory usage, \cite{li:18} proposed Deep Rotation Equivariant Network. They apply rotation transformation on filters rather than feature maps, achieving a speed up of more than 2 times with even less memory overhead. But the methods all need to learn new parameters to achieve rotation equivariant.

\section{Rotation invariance based on LBP operator}
The standard convolutional neural networks do not have the property of rotation invariance. Trained by the upright samples, the performance drops significantly when tested by the rotated images. To solve this problem, we add a regional rotation layer (\emph{RRL}) before the convolutional layers and the fully connected layers. The main idea is that we indirectly achieve rotation invariance by restricting the convolutional features to be rotation equivariant.

\subsection{Local Binary Pattern}
\begin{figure}[t]
	\centering
	\includegraphics[width=0.8\columnwidth]{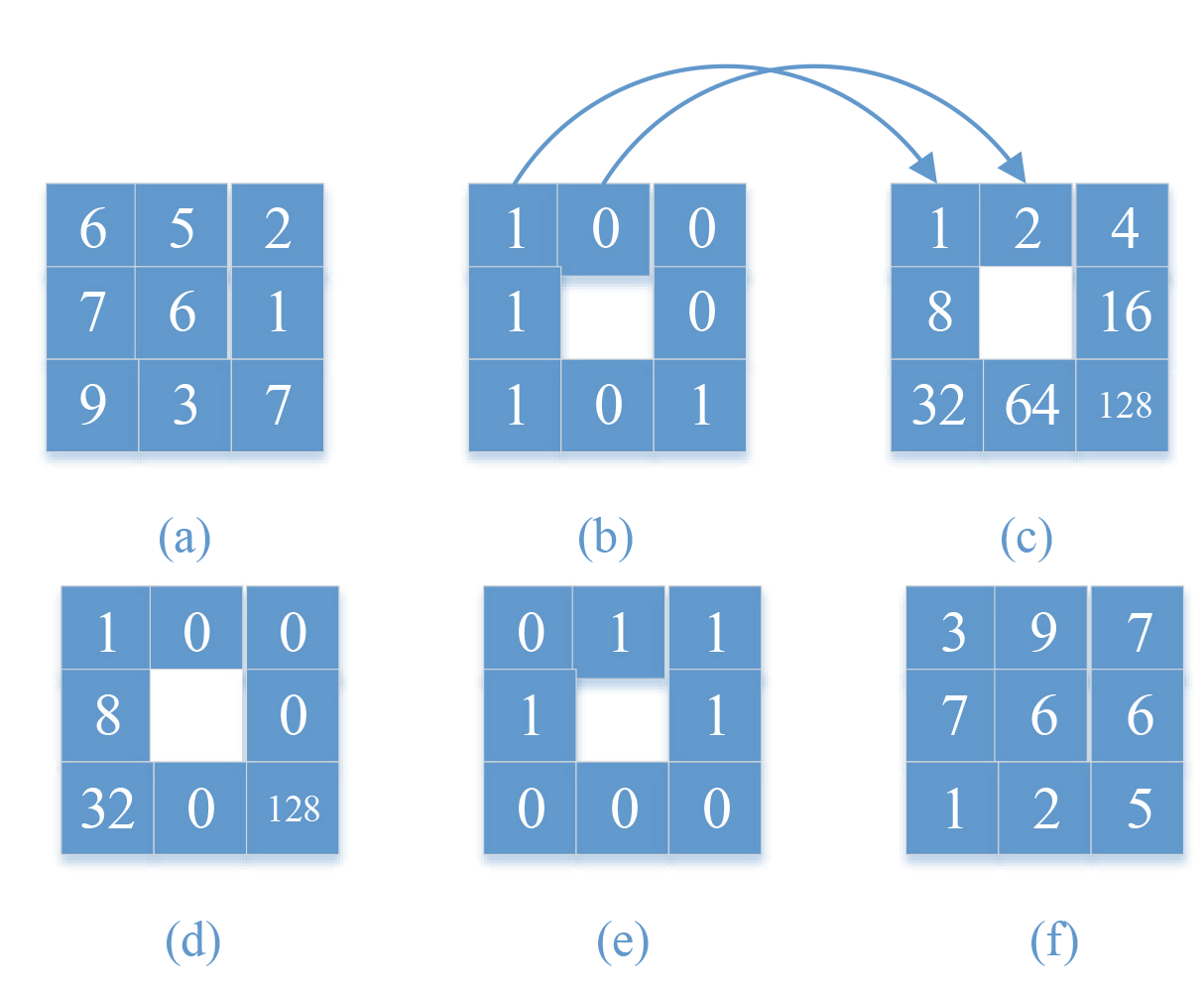} 
	\caption{ Local Binary Pattern. (a) image of size 3×3. The numbers are the gray values; (b) the binary values of the surrounding pixels; (c) the index value of the surrounding pixels; (d) the dot product result of (b) and (c); (e) rotate (b) of 135\textdegree  clockwise to get the minimum LBP; (f) the image window after rotation.}.
	\label{fig1}
\end{figure}

Local Binary Pattern (\emph{LBP}) \cite{o:96} is an operator that describes image texture features. Suppose the window size is $3\times3$, setting the central pixel as the threshold, we can get a binary encoding of the local texture, and convert it to a decimal value. As shown in Figure \ref{fig1}(a), the central point pixel value 6 is used as the comparison reference, then calculate the difference values of the surrounding eight pixels with the central point. If the neighbouring value is less than the central value, the corresponding location is marked as 0, otherwise marked as 1, as shown in Figure \ref{fig1}(b). Taking the upper left corner of the matrix as the starting point, each position is given an index power of 2 according to the flattening and stretching direction of the matrix, as shown in Figure \ref{fig1}(c). The dot production operation is performed between the weight matrix and the binary matrix, as shown in Figure \ref{fig1}(d). Only the values of 1 in the binary matrix are preserved, and the new weight is superimposed. Finally, the surrounding elements of the result matrix are added to form the decimal LBP identifier (in this example 169) of the local texture. A series of LBP feature values are obtained by rotating the surrounding points, and the minimum of these values is selected as the LBP value of the central pixel. In this paper, the points are rotated to the minimum state of LBP, so as to achieve the rotation invariance of angle. In the case, the minimum state is shown as Figure \ref{fig1} (e). That is, the original feature is rotated 135\textdegree clockwise, as shown in Figure \ref{fig1}(f).

\subsection{Regional Rotation Layer (\emph{LBP})}
\emph{LBP} is operated in a window, while \emph{RRL} is operated on the feature maps. The feature maps are sampled one by one in the form of sliding window, and LBP is implemented in each window. So we can rotate the feature maps to the same state even with different input orientations.

\emph{RRL} is usually added before convolutional layer. Here we take the first convolution operation of a three-channel RGB image as an example to illustrate the workflow of \emph{RRL}.

\begin{algorithm}[tb]
	\caption{\emph{RRL} in local windows}
	\label{alg:RRL}
	\textbf{Input}: RGB image sample batch $\{I_1,I_2,\cdots,I_t \cdots\}$\\
	\textbf{Output}: Rotate the feature maps to the same state
	\begin{algorithmic}[1] 
		\STATE Load image $I_t$, $I_t \in \mathbb{R}^{H \times W \times 3}$.
		\STATE Perform \emph{LBP} on each channel to get $V_t,\ V_t\in\mathbb{R}^{F\times F\times(O_H\times O_W\times3)}$, where $F$ is the kernel size, $O_{H(W)}$ is the height/width of the feature map. The sequence of feature maps in the channel has not been changed, shown as process \textcircled{1} in Figure \ref{fig2}.
		\STATE Reshape $V_t$, then concatenate the window features belonging to the same channel into a matrix $I_t^\prime$, $I_t^\prime\in\mathbb{R}^{(F\times O_H)\times(F\times O_W)\times3}$, shown as process \textcircled{2} in Figure \ref{fig2}.
		\STATE Perform a convolution operation with step size $F$ on $I_t^\prime$ and get the output feature $O_t$, $O_t\in\mathbb{R}^{O_H\times O_W\times k^\prime}$, shown as process \textcircled{3} in Figure \ref{fig2}.
	\end{algorithmic}
\end{algorithm}

\begin{figure}[t]
	\centering
	\includegraphics[width=0.8\columnwidth]{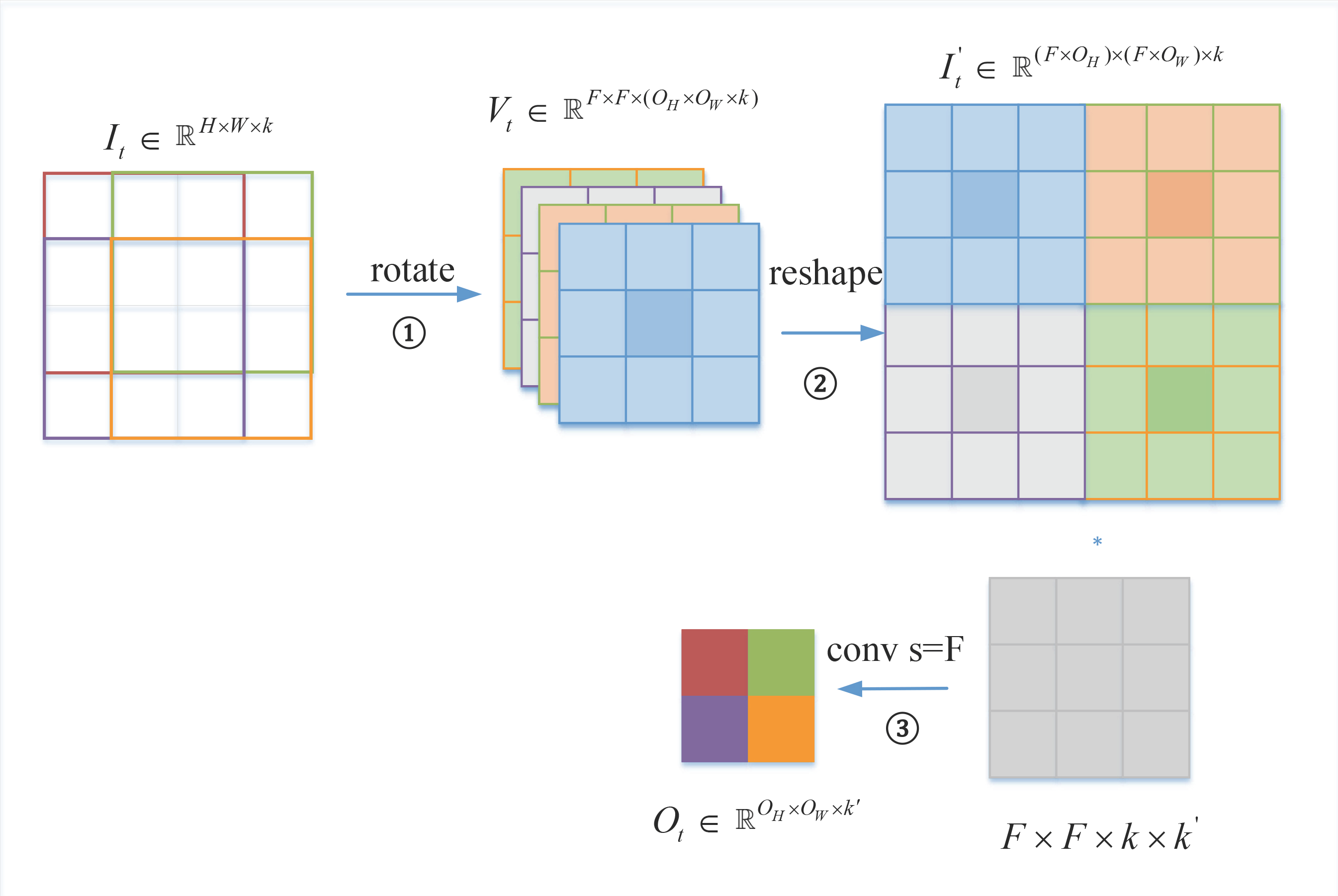} 
	\caption{\emph{RRL} works in a local window $I_t$. Step \textcircled{1}: Perform \emph{LBP} on each channel to rotate feature maps. Step \textcircled{2}: Reshape $V_t$ and concatenate the features into matrix $I_t^\prime$. Step \textcircled{3}: Perform convolution operation on $I_t^\prime$ to get output feature $Q_t$. }.
	\label{fig2}
\end{figure}

\subsection{Rotation Equivariance and Invariance derivation}
Equivariance refers to that, when the transformation can be measured in the output of an operator, the operator and the transformation are equivariant, as shown in eq. \ref{eq1}. 
\begin{equation}
	f(T(x))=T(f(x))
	\label{eq1}
\end{equation}
where $x$ is the input, $T(\bullet)$ is the transformation, $f(\bullet)$ is the operator.

An operator is invariant relative to the transformation, when the influence of the transformation cannot be detected in the output of the operator, shown as eq. \ref{eq2}.
\begin{equation}
f(T(x))=f(x)
\label{eq2}
\end{equation}

In order to achieve rotation invariance of the entire CNN, it is expected that the input features can be rotated uniformly after the convolution layers and before the fully connected layers that perform the classification task.

Local convolution operation is equivariant. In the feature window $w(F \times F)$, when $w$ is rotated of $r$ ($r$ means counterclockwise rotate 90\textdegree, and $r^n$ means counterclockwise rotate n*90\textdegree), if the convolution kernel rotates the same angle $r$, the result is unchanged:$\ f\left(w\right)=L_r[f](rw)$, where $L_r[f]$ indicates that the convolution kernel rotates $r$ counterclockwise. But the local convolution operation is not invariance, when the convolution kernel is unchanged, the results before and after $w$ rotation are different:$\ f\left(w\right)\neq f(rw)$.

 Global convolution operation has neither rotation equivariance nor invariance. When the feature map is rotated, not only the convolution kernel needs to be rotated to the same angle, but the feature output must be rotated with the same angle in the opposite direction, so that the result of the original convolution operation can be kept consistent, as shown in eq. \ref{eq3}.
 \begin{equation}
 f\left(x\right)=r^{-1}L_r[f](rx)
 \label{eq3}
 \end{equation}
 
 From the above analysis, we know that after rotation (only for $r^n$ rotation), the features will still maintain equivariant after layer-by-layer convolution. Therefore, the entire CNN will be rotation invariant if we reversely rotate the feature maps before the fully connected layer.
 
 First, to achieve equivariant of global convolution, the core function $R(x)$ of the algorithm needs to satisfy eq. \ref{eq4}:
 \begin{equation}
 f^F\left[R\left(x\right)\right]=r^{-n}f^F\left[R\left(r^nx\right)\right]
 \label{eq4}
 \end{equation}
 where $f^F\left(x\right)$ is the convolution operation with step size of $F$. In other words, when the filter does not change, the convolution result of the rotated input is equal to that of the non-rotated input through reverse rotating the output. To satisfy eq. \ref{eq4}, the local convolution needs to be invariant: $f\left[R\left(w\right)\right]=f\left[R\left(r^nw\right)\right]$. That is :
 \begin{equation}
 R\left(w\right)=R\left(r^nw\right)
 \label{eq5}
 \end{equation}
 
 Here, we use the core function $R(x)$, which is named \emph{RRL} module, to make the window convolution invariant, and achieve rotation invariant of the CNN. \emph{RRL}s’ position is before each convolutional layer and after the last convolutional layer with the step size of $F$. In particular, for the last \emph{RRL}, the global feature maps $x$ are treated as a local window $w$, and satisfies $R\left(x\right)=R\left(r^nx\right)$. Because the activation function, BN layer and pooling layer are rotation equivariant and they do not affect the final result, they are not discussed here.
 
 \subsection{Integrate RRL with CNN}
 Each \emph{RRL} $R_i$ is embedded before each conv layer $f_i$. Suppose that the original feature $x$  and the rotated 90\textdegree feature $rx$ are fed into $R_1$ respectively.
 
 After $R_1$ and $f_1$, we have:
 \begin{displaymath}
 \begin{split}
& x\rightarrow f_1^{F_1}\left[R_1\left(x\right)\right]\\
& rx\rightarrow f_1^{F_1}\left[R_1\left(rx\right)\right]\\
 \therefore & f_1^{F_1}\left[R_1\left(x\right)\right]=r^{-1}f_1^{F_1}\left[R_1\left(rx\right)\right]
 \end{split}
 \end{displaymath}
After $R_2$ and $f_2$, we have: 
\begin{displaymath}
\begin{split}
& f_1^{F_1}\left[R_1\left(x\right)\right]\rightarrow f_2^{F_2}\left[R_2\left[f_1^{F_1}\left[R_1\left(x\right)\right]\right]\right]\\
& f_1^{F_1}\left[R_1\left(rx\right)\right]\rightarrow f_2^{F_2}\left[R_2\left[f_1^{F_1}\left[R_1\left(rx\right)\right]\right]\right]\\
\therefore & { f}_2^{F_2}\left[R_2\left[f_1^{F_1}\left[R_1\left(x\right)\right]\right]\right]=r^{-1}f_2^{F_2}\left[R_2\left[f_1^{F_1}\left[R_1\left(rx\right)\right]\right]\right]
\end{split}
\end{displaymath} 
So, we have:
\begin{equation}
\begin{split}
R_{n+1}\left[f_n^{F_n}\left[\ldots\left[R_2\left[f_1^{F_1}\left[R_1\left(x\right)\right]\right]\right]\right]\right]\\=R_{n+1}\left[f_n^{F_n}\left[\ldots\left[R_2\left[f_1^{F_1}\left[R_1\left(rx\right)\right]\right]\right]\right]\right] 
\label{eq6}
\end{split}
\end{equation}
The conclusion can be extended to other CNNs. When \emph{RRL} is added in the right position, the rotation invariance of the model can be achieved.

\section{Experiments}
\subsection{Image classification based on LeNet-5}
\subsubsection{Dataset and LeNet-5}
\begin{figure}[t]
	\centering
	\includegraphics[width=0.9\columnwidth]{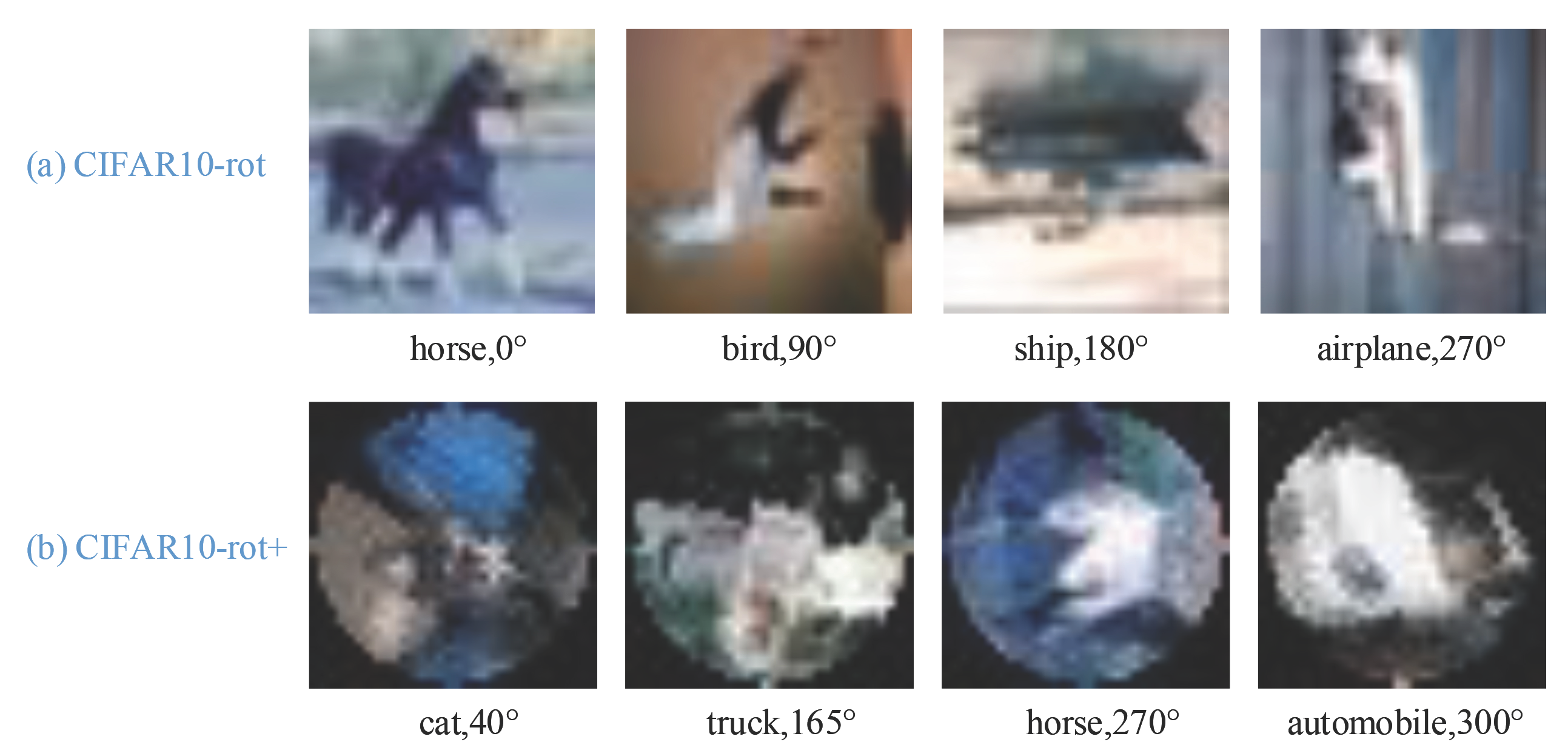} 
	\caption{Examples of CIFAR10-rot and CIFAR10-rot+}.
	\label{fig4}
\end{figure}

CIFAR-10 \cite{k:09} is used in our experiment. The dataset was proposed by krizhevsky in 2009. It contains 60000 $32\times32$ colour images, belonging to 10 categories. There are 50000 images in training set (5000 in each category) and 10000 images in test set (1000 in each category). The images rotated in the first ways are call CIFAR10-rot (namely $\theta\in\{0,90,180,270\}$\textdegree and in the second way are called CIFAR10-rot+ ( $\theta\in[0,360)$\textdegree), as shown in Figure \ref{fig4}. In order to ensure that the effective content area of the image is fixed, the largest inscribed circle of the square image is selected. Only the inner area of the circle has original image pixels, and the outer area of the circle is filled with black, shown as Figure \ref{fig4} (b).

LeNet-5 \cite{l:98} is one of the earliest CNNs. It has two convolutional layers and three fully connected layers, so three RRLs are plugged in. Keeping the convolutional layers and fully connected layers unchanged as \cite{l:98}, the three RRLs are inserted in front of conv1, conv2 and after conv2, respectively.
\subsubsection{Experimental result and analysis}
Table \ref{table1} shows the test accuracy of LeNet-5 on CIFAR-10 with and without \emph{RRL}. The second column is trained by the original training set (without rotation images) and tested by CIFAR10-rot. The third column is trained by original training set (without rotation) and tested by CIFAR10-rot+ data set. The fourth and fifth columns are trained and tested by CIFAR10-rot and CIFAR10-rot + datasets respectively.

From Table \ref{table1} we can find that:

1) Keeping the original CNN structure and adding only the RRLs can improve the recognition accuracy of rotating images greatly;

2) Trained with the augmented data, the accuracy of improved network decreases. It implies that LeNet-5 cannot provide more convolution kernels to learn the same patterns with different directions, so it reduces generalization;

3) Without RRLs, the accuracy of recognition can be improved a bit by using data augmentation, but the training cost increases and the problem is not solved essentially.

\begin{table}[t]
	\centering
	\begin{tabular}{l|p{1cm}|p{1.2cm}|p{1cm}|p{1.2cm}}
		\hline
		Training Data & CIFAR-10 & CIFAR-10 & CIFA10-10-rot & CIFAR-10-rot+ \\
		\hline
		Testing Data & CIFAR-10-rot & CIFAR-10-rot+ & CIFAR-10-rot & CIFAR-10-rot+\\
		\hline
		LeNet-5 & 33.2 & 18.2 & 38.7 & 25.4 \\
		\hline
		LeNet-5+\emph{RRL} & 71.3 & 52.8 & 70.9 & 49.1 \\
		\hline
	\end{tabular}
	\caption{Comparison of accuracy (\%) on LeNet-5 with and without \emph{RRL}.}
	\label{table1}
\end{table}

\begin{figure}[t]
	\centering
	\includegraphics[width=0.9\columnwidth]{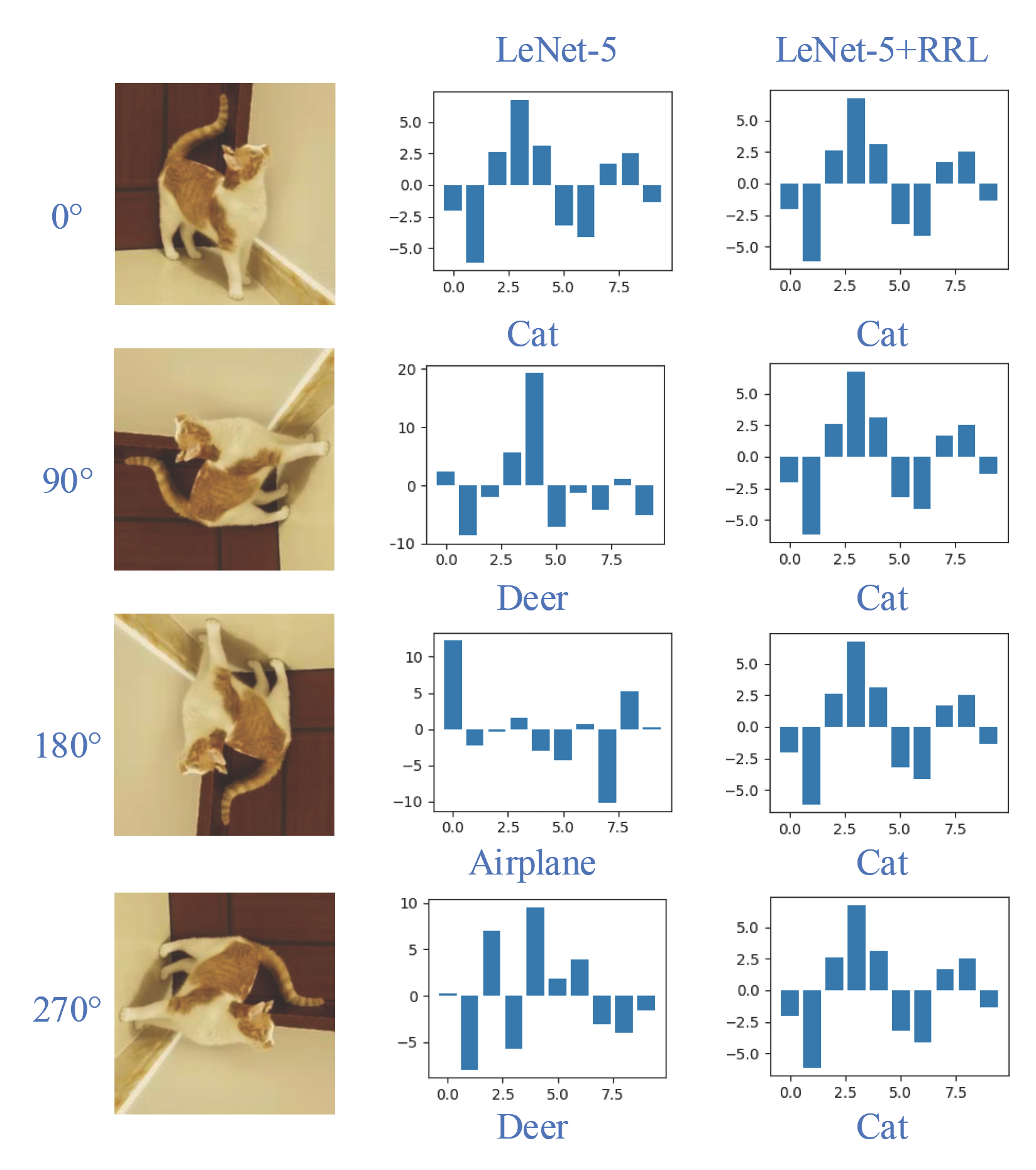} 
	\caption{Feature distributions of LeNet-5 with and without \emph{RRL}s, input with different rotation angles }.
	\label{fig5}
\end{figure}

In order to analyze the role of \emph{RRL} more intuitively, the feature maps are visualized in Figure \ref{fig5}. In Figure \ref{fig5}, the left columns are the input images. The middle columns are the output feature of the last layer of the original LeNet-5 network. Except the upright image can be correctly classified, the other three cases are misidentified. The right columns are the output features of the last layer of LeNet-5+RRL network, whose features do not change with the rotation angle, and all predict the correct category. It shows that with the RRLs, the same features are extracted from the images with different angles, and the coding invariance is realized.

\begin{figure}[t]
	\centering
	\includegraphics[width=0.9\columnwidth]{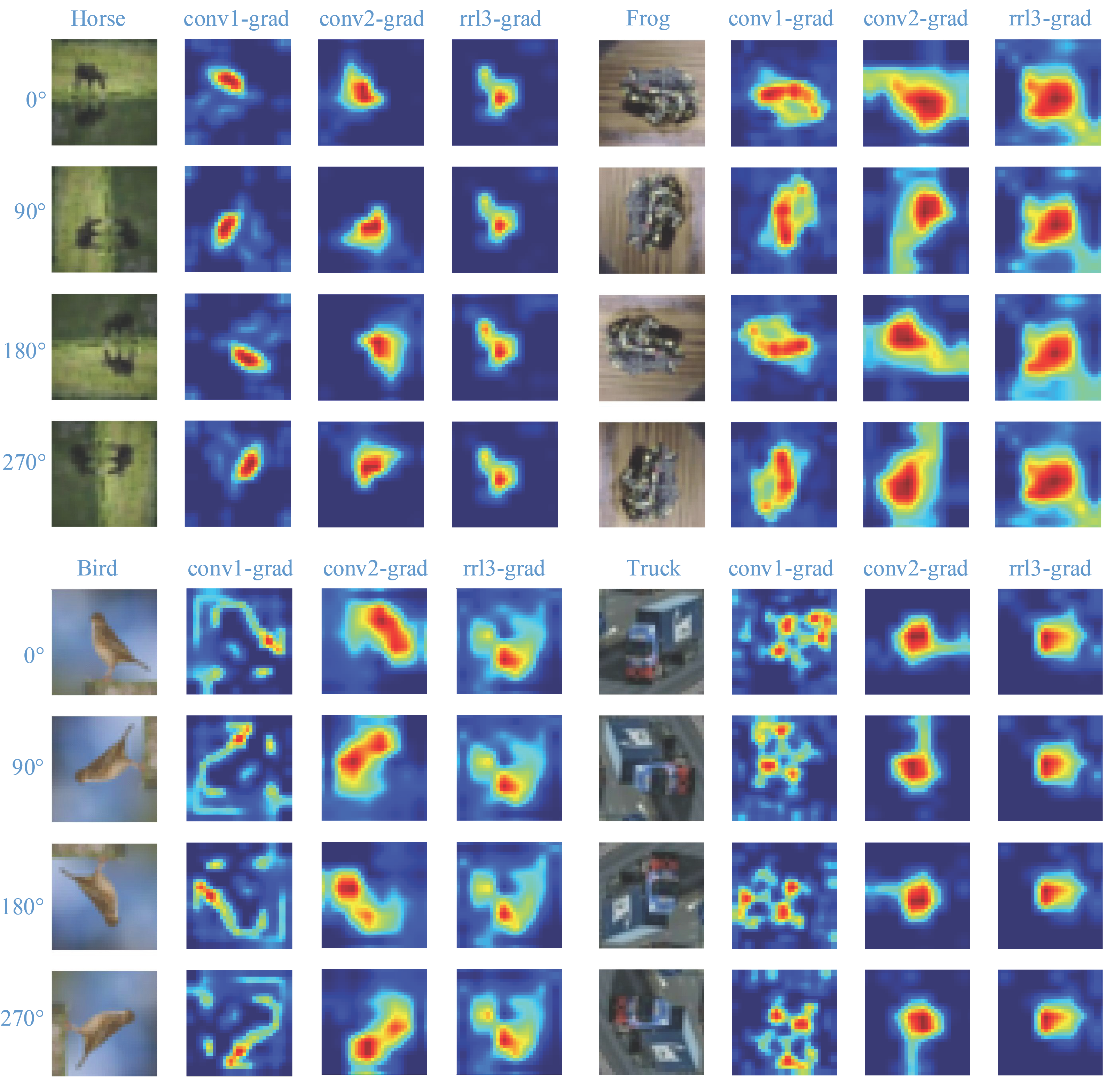} 
	\caption{Visualization of three \emph{RRL}s output heatmaps in LeNet-5 + \emph{RRL} }.
	\label{fig6}
\end{figure}

The visualization results of Grad-CAM\cite{s:17grad} are shown in Figure \ref{fig6}. The input images are rotated $\theta \in \{0,90,180,270\}$\textdegree. Conv1-grad, Conv2-grad and RRL3-grad are the heatmaps obtained by gradient calculation of the feature maps after the first layer, the second layer of convolution and the last regional rotation layer, respectively. It can be seen that before the last regional rotation, the features are direction dependent, and the focus position of the model still changes with the rotation angle of input. At this stage, the network is only rotation equivariant. After RRL3, the neural network completes the invariance coding of rotation, and the features shown by RRL3-grad hardly change with the rotation.

\subsection{Image classification based on ResNet-18}
ResNet-18 \cite{h:16} was proposed in 2016, and it consists of 17 convolution layers and a fully connected layer. The core component of ResNet is the residual module, which consists of two consecutive convolution layers and a skip connection.

Table 2 shows the comparison of the accuracy of ResNet-18 with and without \emph{RRL}. It can be seen from Table \ref{table2} that the effect of data augmentation is not significant with \emph{RRL}. No matter whether the input sample is rotated or not, as long as the sample itself remains unchanged, the local features will remain unchanged after \emph{RRL}s.

\begin{table}[t]
	\centering
	\begin{tabular}{l|p{1cm}|p{1.2cm}|p{1cm}|p{1.2cm}}
		\hline
		Training Data & CIFAR-10 & CIFAR-10 & CIFA10-10-rot & CIFAR-10-rot+ \\
		\hline
		Testing Data & CIFAR-10-rot & CIFAR-10-rot+ & CIFAR-10-rot & CIFAR-10-rot+\\
		\hline
		ResNet-18 & 46.5 & 38.7 & 73.6 & 58.7 \\
		\hline
		ResNet-18+\emph{RRL} & 75.0 & 65.3 & 77.9 &63.1 \\
		\hline
	\end{tabular}
	\caption{Comparison of accuracy (\%) ResNet-18 with and without \emph{RRL}.}
	\label{table2}
\end{table}

Comparison with other methods on CIFAR-10 is shown in Table \ref{table3}. We can see that ResNet-18+\emph{RRL} has obtained high accuracy on both data sets. It shows that \emph{RRL} can help the original CNN to improve the encoding ability without increasing the parameters and model complexity, and obtain stronger generalization ability.

From table \ref{table3}, we can also find that ResNet + \emph{RRL} improves performance less than LeNet + \emph{RRL} does (28.5\% vs 38.1\% on CIFAR10-rot, 26.6\% vs 34.1\% on CIFAR10-rot+). \emph{LBP} operator tends to rotate the brighter texture of the image to the left part of the window. We can guess that with the restriction of \emph{RRL}, the obtained features tend to be similar and reduce the diversity of features. After the training data are enhanced, the gap between the two is also significantly smaller (improve 4.3\% on CIFAR10-rot and 4.4\% on CIFAR10-rot+ respectively). For rotated images, the traditional convolutional network will specially customize the filter for each direction of the same texture. However, due to the constraint of \emph{RRL}, even if the input data are more diverse, the feature types with little change in content will not increase significantly. Therefore, when the model is deepened, the tradition neural networks will improve more than that of networks with \emph{RRL}.

\begin{table}[t]
	\centering
	\begin{tabular}{p{3cm}|c|c}
		\hline
		Training Data & \multicolumn{2}{c}{CIFAR-10} \\
		\hline
		Testing Data & CIFAR-10-rot & CIFAR-10-rot+ \\
		\hline
		LeNet-5 & 33.2 & 18.2 \\
		\hline
		LeNet-5+\emph{RRL} & 71.3 & 52.8 \\
		\hline
		ResNet-18 & 46.5 & 38.7 \\
		\hline
		ResNet-18+\emph{RRL} & 75.0 & 65.3 \\
		\hline
		CyResNet56-P \cite{c:15pl} & - & 61.3 \\
		\hline
		PR\_RF\_1 \cite{f:18r} & - & 44.1 \\
		\hline
		ORN \cite{z:17} & 60.9 & 40.7 \\
		\hline
	\end{tabular}
	\caption{Comparison of accuracy (\%) with other methods on CIFAR-10.}
	\label{table3}
\end{table}

Even with data augmentation, the rotation invariance of conventional convolution network is still not as good as plugged with \emph{RRL}. However, it can be predicted that with the deepening of the network, the rising trend of accuracy with \emph{RRL} will slow down. Therefore, the algorithm is suitable for shallow or medium neural networks or limited training samples and limited computing resources.

\begin{figure}[t]
	\centering
	\includegraphics[width=0.9\columnwidth]{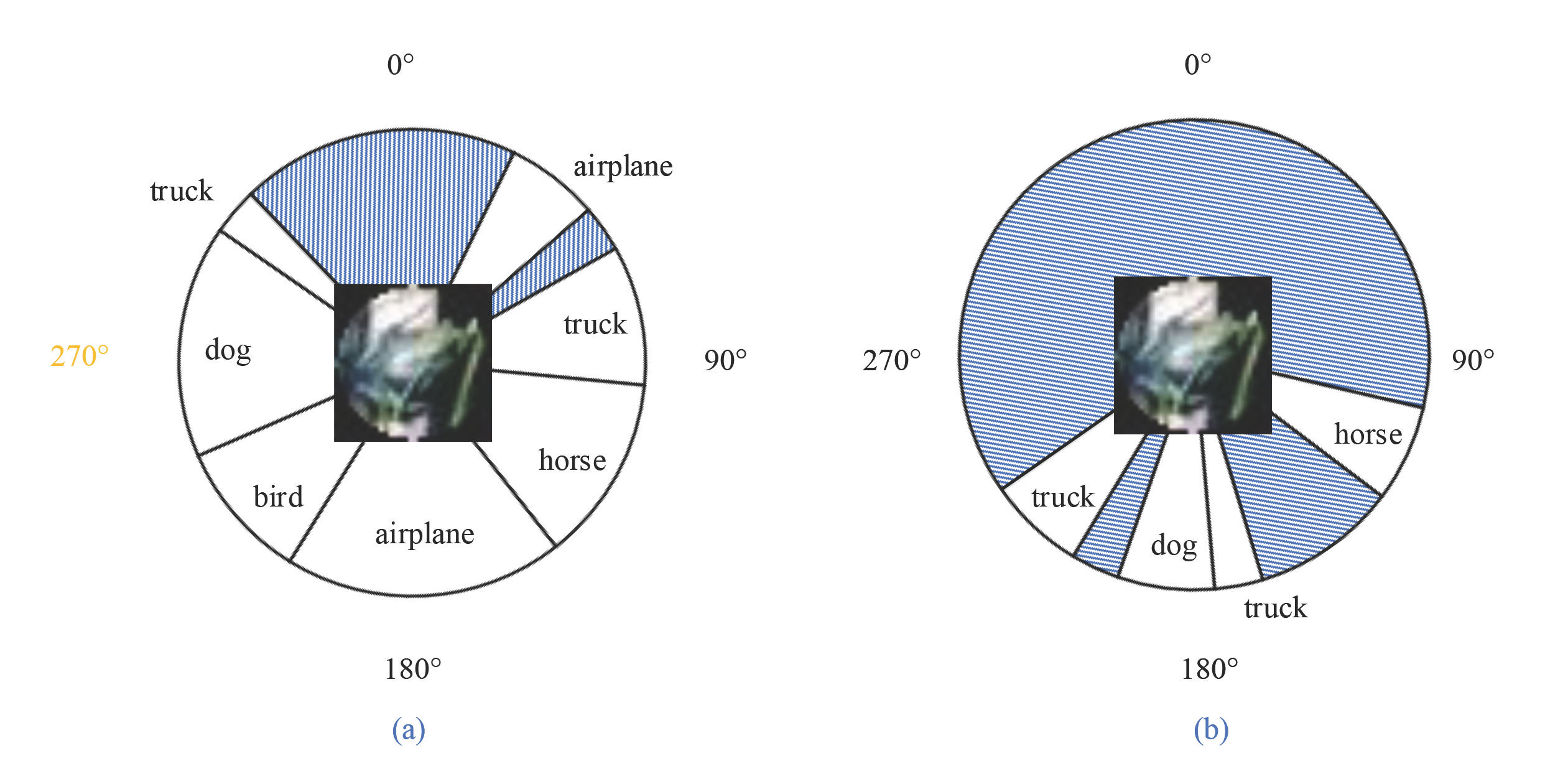} 
	\caption{Recognition results with arbitrary rotation angles. (a)ResNet-18, (b)ResNet-18 + \emph{RRL} }.
	\label{fig8}
\end{figure}

Figure \ref{fig8} shows the classification results of the same image at different rotation angles with and without \emph{RRL}s. The blue sections mean the angle range of correctly classifying "frogs". The image is rotated every 12\textdegree, thus there are 30 rotation angle sections. Figure \ref{fig8}(a) shows the classification output of the model without \emph{RRL}s. When the rotation angle is between $\theta\in[-36,24]$\textdegree  or $\theta\in[36,60]$\textdegree, it can be classified correctly. In other states, different prediction results will be obtained with different angles. Figure \ref{fig8}(b) shows the classification output of the improved model. The blue area is larger than that of (a), indicating that the rotation layers makes the model more insensitive to the input rotation angle.

\subsection{Plankton Recognition based on ResNet-44}
\begin{figure}[t]
	\centering
	\includegraphics[width=0.9\columnwidth]{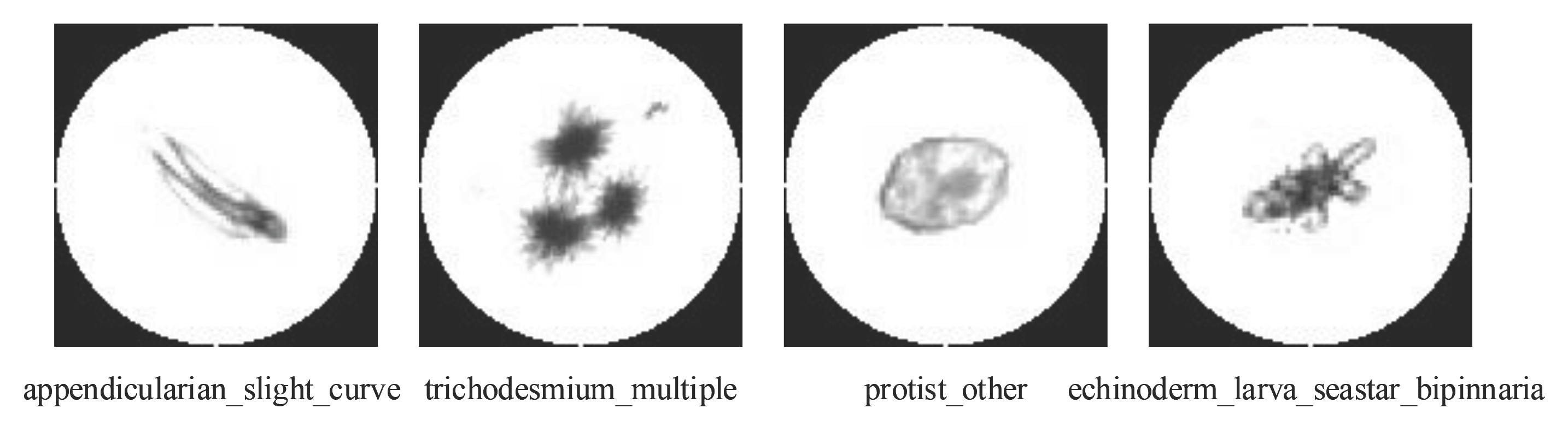} 
	\caption{Sample images in Plankton dataset }.
	\label{fig9}
\end{figure}

The plankton dataset\cite{c:15pl} consists of 30,336 gray images of different sizes, which are unevenly divided into 121 categories, corresponding to different kinds of plankton. There are 27,299 images in training set and 3,037 images in testing set. In order to unify the sample number of each category, the data are augmented for the categories with a small number of images. Finally, each category in the training set contains 2000 images and each category in the testing set contains 100 images. Each sample resizes to $50\times50$, then pads with white background to $64\times64$, and finally take its maximum inscribed circle to ensure that the image is in the centre of the image. Figure \ref{fig9} shows the results of data processing and their categories.

Each image contains a single organism, which may be in any direction in three-dimensional space due to ignoring the small influence of gravity. And the ocean is full of debris particles, so there will inevitably be some noise in the image. The existence of unknown categories requires the model to deal with unrecognized objects, so it is necessary to classify those plankton with large shape differences into the same category. The above factors make the classification more difficult.

The standard ResNet-44 consists of 43 convolutional layers and a fully connected layer.A regional rotation layer is added in front of all convolutional layers.

After 100,000 epochs of training on the training set, the final plankton classification model is obtained. In order to compare the effect of regional rotation layer, the original ResNet-44 and the ResNet-44 + RRL model after adding regional rotation layer are trained and tested respectively.

\begin{figure}[t]
	\centering
	\includegraphics[width=0.9\columnwidth]{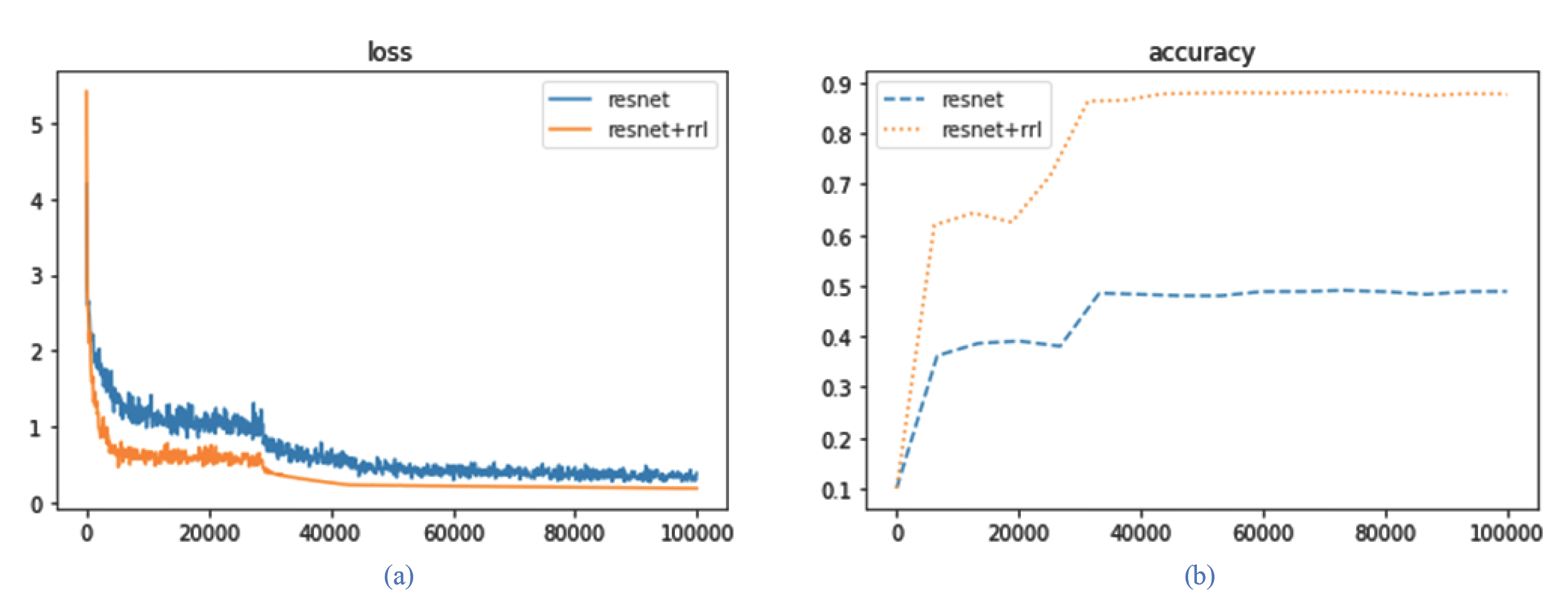} 
	\caption{Comparison of loss and accuracy with and without \emph{RRL} by ResNet-44. The blue curve is the orginal ResNet-44 model, and the orange curve is ResNet-44+\emph{RRL}.}.
	\label{fig10}
\end{figure}

Figure \ref{fig10}(a) shows the loss curves of the two algorithms on the training set. The orange curve is ResNet-44 + RRL model, and the blue curve is the original ResNet-44 model. It can be seen that the orange curve is always smaller than the blue curve, that is, the regional rotation layer makes the model error smaller. Figure \ref{fig10}(b) shows the accuracy curves of the two algorithms on the testing set. Obviously, after adding the regional rotation layer, the error of the training set is reduced, and there is no over-fitting, and the performance is improved. Table \ref{table4} shows the running results of applying the two models to the real test set without published labels. The lower the score means the model performs better. The performance of the dataset shows that without increasing the model parameters, the convolutional neural network can have stronger generalization by adding a regional rotation layer, and give the neural network the ability to capture global and local rotation.

\begin{table}[t]
	\centering
	\begin{tabular}{c|c}
		\hline
		Model & Multi-class Score \\
		\hline
		ResNet-44 & 3.67862 \\
		\hline
		ResNet-44+\emph{RRL} & 2.18777\\
		\hline
	\end{tabular}
	\caption{ Comparison of loss scores of real testing sets on Plankton dataset.}
	\label{table4}
\end{table}

\subsection{Object detection based on MobileNet-tiny-yolov3}

MobileNet-tiny-yolov3 is selected as the basic network. Compared with the darknet53 with residual as the main structure, mobilenet can achieve a better balance in terms of calculation, storage space and accuracy. Using the pruned tiny yolov3, the model is smaller and has more advantages when the computing resources are limited, and the fast detection speed also makes tiny yolov3 more cost-effective and easier to be applied in practice.

\subsubsection{Rotation transformation of coordinate}
\begin{figure}[t]
	\centering
	\includegraphics[width=0.9\columnwidth]{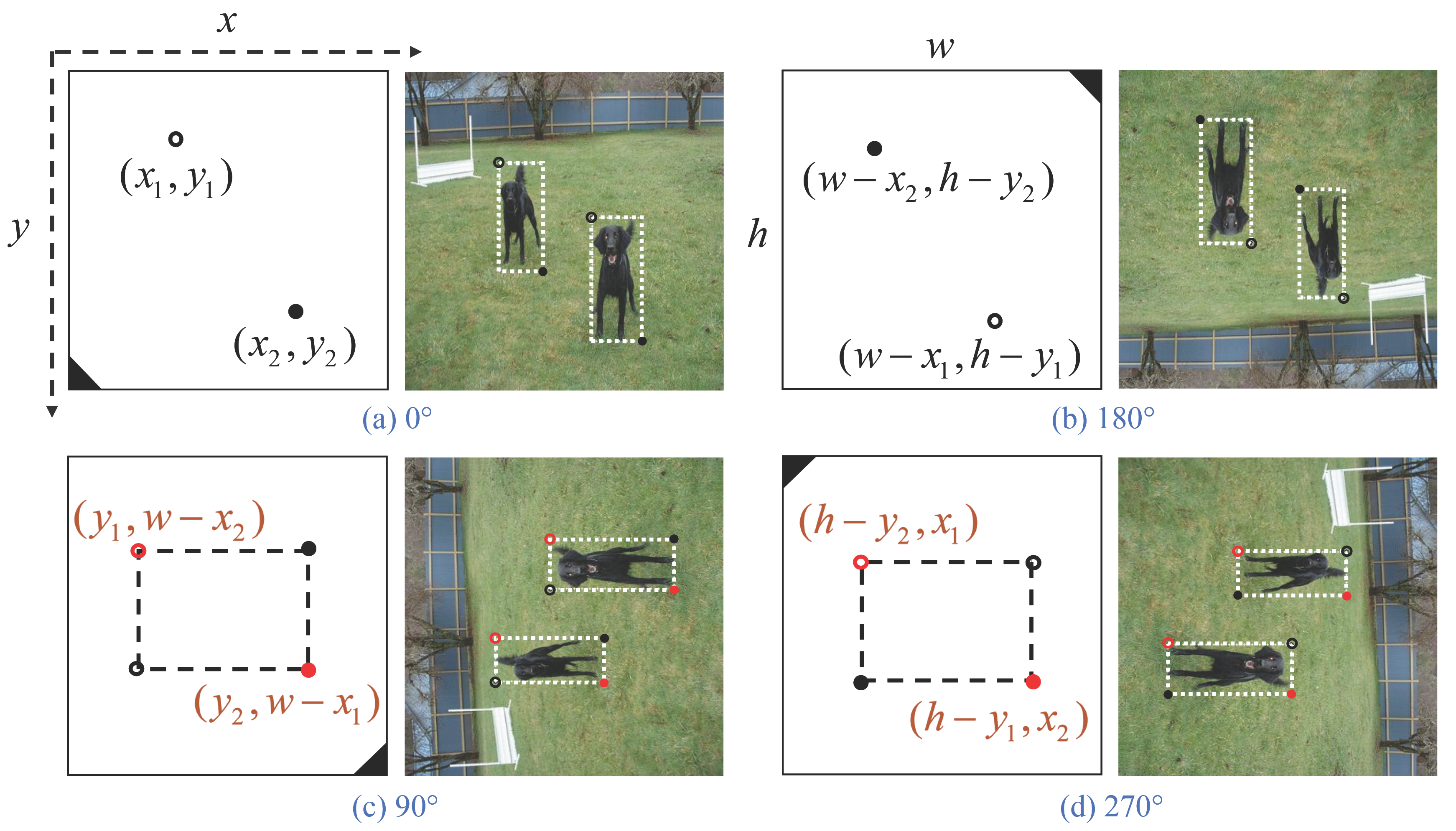} 
	\caption{Coordinate transformation}.
	\label{fig11}
\end{figure}

In the object detection task, the coordinates of the upper left corner and the lower right corner of the target bounding boxes are usually provided as labels, so the location changes with the rotation of the target. The corresponding coordinate labels need to be recalculated. Here we only consider four rotation angles, $\theta \in \{0,90,180,270\}$ \textdegree. As shown in Figure \ref{fig11}. There is a rectangular box surrounding the target object, which is defined by the upper left coordinate $(x_1, y_1)$ and the lower right coordinate $(x_2, y_2)$. When the image rotates 90\textdegree counterclockwise, the point $(x_1, y_1)$ is transformed into $(y_1, w-x_1)$, the point $(x_2, y_2)$ is transformed into $(y_2, w-x_2)$, and it represents the points in the lower left corner and upper right corner of the rectangular box respectively. The final coordinate label becomes a red hollow point $(y_1, w-x_1)$ and a red solid point $(y_2, w-x_2)$. Similarly, (b) and (d) represent the position label when rotating 180\textdegree and 270\textdegree counterclockwise.

\subsubsection{Dataset}
Pascal VOC dataset contains 20 categories. The dataset has been widely used in object detection, semantic segmentation and classification tasks, and as a common test benchmark. VOC 2007 and VOC 2012 are used in this experiment. Finally, 16,551 images and 40,058 objects are used in training, 4,952 images and 12,032 objects are used in testing.

\begin{figure}[t]
	\centering
	\includegraphics[width=0.9\columnwidth]{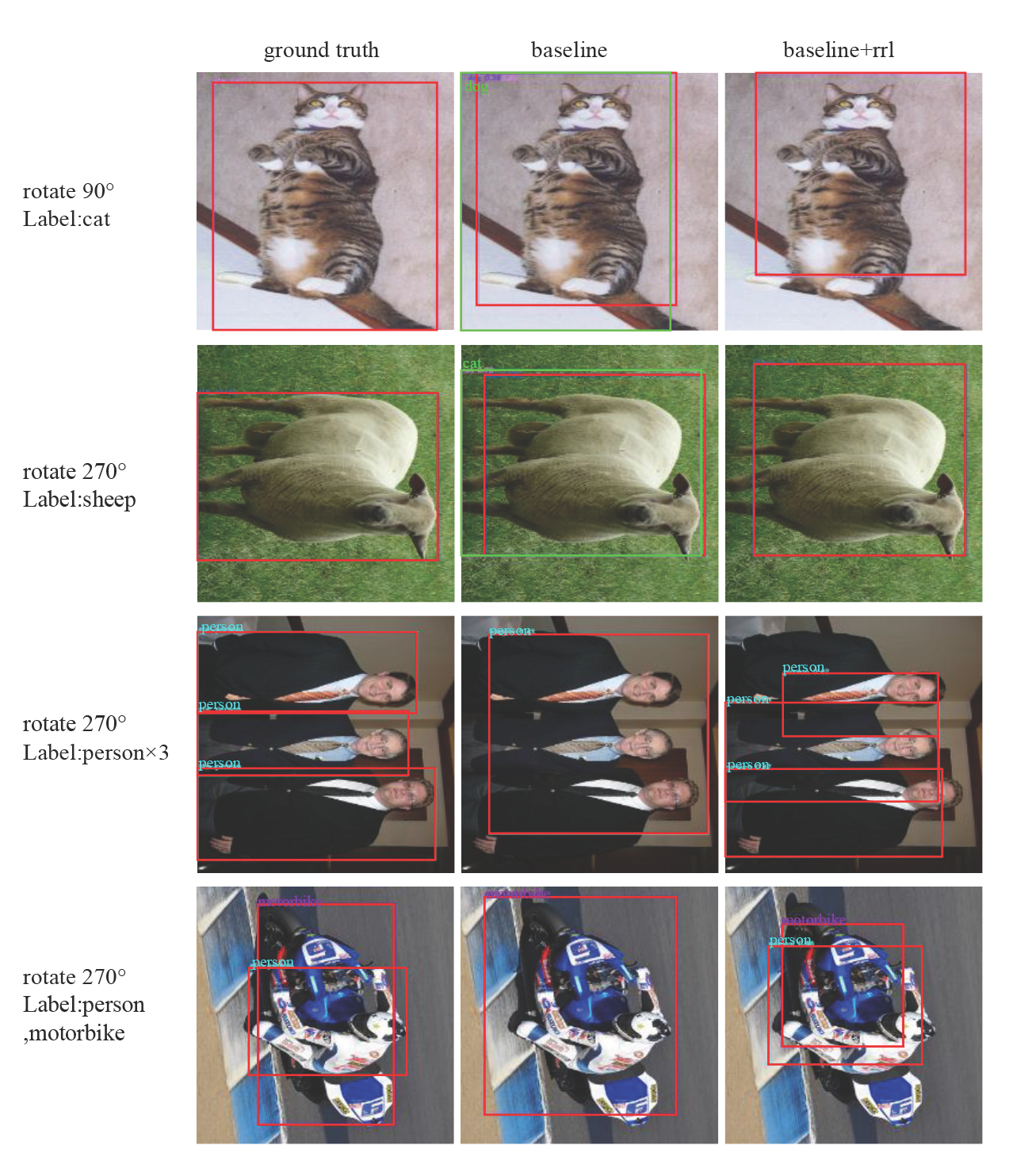} 
	\caption{Examples of detection effect before and after model improvement}.
	\label{fig12}
\end{figure}

\subsubsection{Experimental results and analysis}

The detection effect of MobileNet-tiny-yolov3 with and without \emph{RRL} are shown in Figure \ref{fig12}. Both models are trained with upright images. The first column shows the groundtruth after rotating and scaling the original image, the second and third column are the testing results on the basic model, and adding \emph{RRL}s. For the top two rows, the image contains only one label, but the basic model outputs two prediction boxes, one of which does not belong to the correct category, as shown in the green box. It can be seen that the basic model has some recognition ability for rotating images, and will be misled into other wrong categories. The output by the improved model is quite similar with real label. The bottom two rows contain multiple labels, and they all overlap to some extent. The output of the basic model contains only one target and all objects. It shows that the model can only be roughly positioned, and can no longer be finely divided. The improved model can detect each object with some location errors. IoU threshold is set to 0.5, trained with upright pictures and tested with rotating pictures. The mAP of the two models is shown in the table \ref{table5}.

\begin{table}[t]
	\centering
	\begin{tabular}{c|c}
		\hline
		Model & mAP \\
		\hline
		MobieNet-tiny-yolov3 & 43.76\% \\
		\hline
		MobieNet-tiny-yolov3+\emph{RRL} & 61.78\% \\
		\hline
	\end{tabular}
	\caption{ mAP of MobileNet-tiny-yolov3 on Pascal VOC dataset. Trained by upright images and tested by rotated images.}
	\label{table5}
\end{table}

\section{Conclusion}
This paper proposes a regional rotation layer (\emph{RRL}) to help CNNs to learn rotation invariant features. By data augmentation, CNN needs to train more filters for each change in the sample, which leads to the increase of the number of parameters. So it is important to balance the network size and the data size. In this paper, \emph{LBP} operator is used to encode the local region so that it has the same local features before and after rotation. Thus, when the input changes, the local features remain the same. Then \emph{RRL} is integrated with LeNet-5, ResNet-18 and tiny-yolov3, which verifies the effectiveness of the method. Experimental results are analysed in detail, the applicable scenarios and shortcomings of the method are presented.

\section{Acknoledgement}

This project is supported by the fund of Science and Technology of Sichuan Province (No.2021YFG0330). Thanks for Shuyu Zhang, Yiting Wang.


\bibliography{aaai22.bib}
\end{document}